\documentclass[letterpaper, 10 pt, conference]{ieeeconf}  

\IEEEoverridecommandlockouts                              

\usepackage{graphicx}

\title{\LARGE \bf
Thoracic Surgery Video Analysis for Surgical Phase Recognition}

\author{Syed Abdul Mateen$^{1}$ \qquad Niharika Malvia$^{1}$ \qquad Syed Abdul Khader$^{2}$ \qquad Danny Wang$^{3}$ \qquad Deepti Srinivasan$^{4}$ \\\\
Chi-Fu Jeffrey Yang$^{3}$ \qquad Lana Schumacher$^{4}$ \qquad Sandeep Manjanna$^{2}$
\thanks{$^{1}$Indian Institute of Technology Palakkad, Palakkad, India.}%
\thanks{$^{2}$Plaksha University, Mohali, India.}
\thanks{$^{3}$Massachusetts General Hospital, Boston, USA.}
\thanks{$^{4}$Tufts University, Medford, USA.}
\thanks{{Conrrespondance: \tt\small msandeep.sjce@gmail.com}}%
}

\begin{document}

\maketitle
\thispagestyle{empty}
\pagestyle{empty}


\begin{abstract}

This paper presents an approach for surgical phase recognition using video data, aiming to provide a comprehensive understanding of surgical procedures for automated workflow analysis. The advent of robotic surgery, digitized operating rooms, and the generation of vast amounts of data have opened doors for the application of machine learning and computer vision in the analysis of surgical videos. Among these advancements, Surgical Phase Recognition(SPR) stands out as an emerging technology that has the potential to recognize and assess the ongoing surgical scenario, summarize the surgery, evaluate surgical skills, offer surgical decision support, and facilitate medical training. In this paper, we analyse and evaluate both frame-based and video clipping-based phase recognition on thoracic surgery dataset consisting of 11 classes of phases. Specifically, we utilize ImageNet ViT for image-based classification and VideoMAE as the baseline model for video-based classification. We show that Masked Video Distillation(MVD) exhibits superior performance, achieving a top-1 accuracy of $72.9\%$, compared to $52.31\%$ achieved by ImageNet ViT. These findings underscore the efficacy of video-based classifiers over their image-based counterparts in surgical phase recognition tasks.

\end{abstract}

\section{INTRODUCTION}

Robot-assisted surgery has become increasingly adopted in recent years and has expanded the scope of treatment options for patients~\cite{mederos2022trends}. The advent of robotic surgery, digitized operating rooms, and the generation of vast amounts of data have opened doors for the application of machine learning and computer vision in the analysis of surgical videos. Within this field of surgical data science, automating data analysis is crucial to simplify complexity and maximize data utility, enabling new opportunities. For example, automating intraoperative assistance and cognitive guidance for surgeons in real-time offers potential benefits, as does providing automated and enhanced feedback for trainees. This is especially useful for fields like thoracic surgery, where several reported cases of intraoperative catastrophes still occur and require conversion to an open thoracotomy~\cite{servais2022conversion}. Moreover, autonomously analyzing surgical data has the potential to optimize entire surgical workflows.

Among these advancements, Surgical Phase Recognition(SPR) stands out as an emerging technology that has the potential to recognize and assess the ongoing surgical scenario, summarize the surgery, evaluate surgical skills, offer surgical decision support, and facilitate medical training. In this paper, we analyse and evaluate both frame-based and video clipping-based phase recognition on thoracic surgery dataset consisting of 11 classes of phases. Specifically, we apply ImageNet ViT for image-based classification, and VideoMAE and Masked Video Distillation(MVD) for video-based classification. We show that video-based SPR performs significantly better than image-based SPR, achieving a top-1 accuracy of $72.9\%$, compared to $52.31\%$ achieved by ImageNet ViT.

\section{Methodology}

We employ three models for SPR: ImageNet-pretrained Vision Transformer (ViT)~\cite{beyer2022better}, Video Masked Autoencoder (VideoMAE)~\cite{tong2022videomae}, and Masked Video Distillation (MVD)~\cite{wang2023masked}. While ImageNet ViT is an image-based model, the latter two are video-based models specifically designed for video understanding tasks. For VideoMAE and MVD, we utilized a ViT-L backbone model that was pretrained on the Kinetics-400 dataset~\cite{kay2017kinetics}. This dataset is a large-scale video dataset commonly used for pretraining video models. While both models employ masked feature modeling and utilize an encoder-decoder transformer architecture, MVD introduces an additional feature. This feature involves the use of transfer learning to transfer knowledge from pretrained image and video teacher models to a designated student model, referred to as the student encoder. By leveraging the knowledge learned by these models, we aim to enhance the performance of our models on the surgical phase classification task.

\vspace{-1em}
\begin{figure}[h]
    \centering
    \includegraphics[width=1\linewidth]{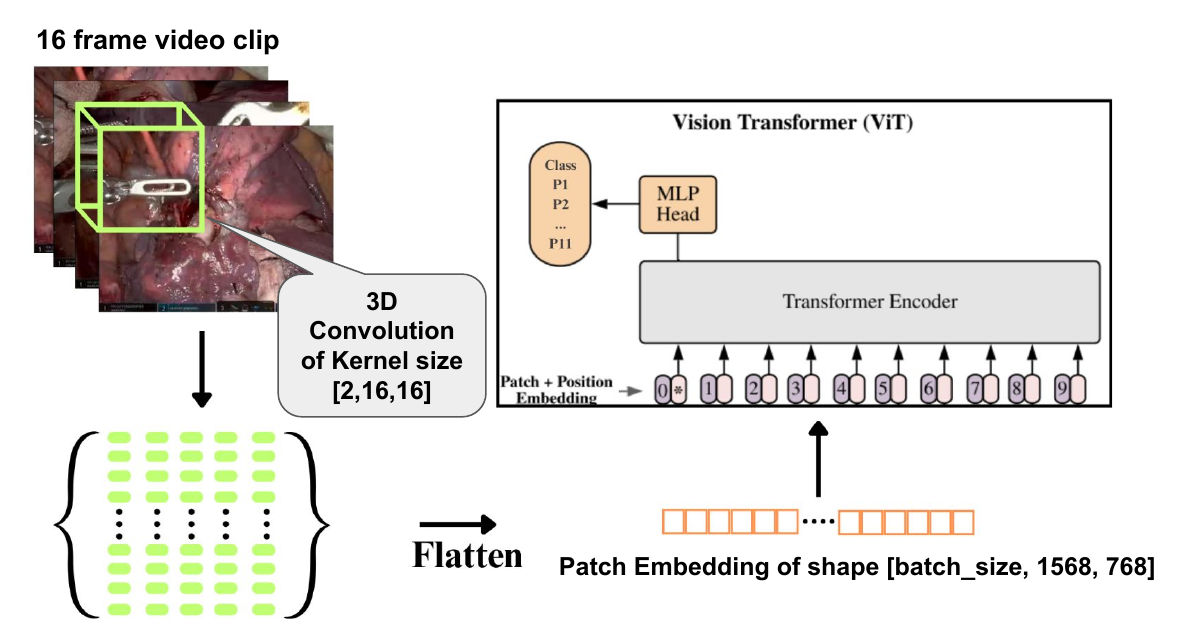}
    \caption{The input video, comprising 16 frames, undergoes a 3D patch embedding. Then, positional embedding incorporates spatial information, followed by the input's passage through the transformer encoder used in Vision transformer~\cite{dosovitskiy2021image} for feature extraction.}\vspace{-1em}
    \label{fig:overview}
\end{figure}

To adapt the pretrained models to our specific task, we fine-tuned VideoMAE and MVD for 100 epochs by monitoring the Top-1 Accuracy metric to compare the performance of checkpoints between epochs. This facilitated the selection of the best-performing model for each architecture. Fig.~\ref{fig:overview} illustrates an overview of the architecture used. To evaluate the generalization capability of our models, we split the 17 patient dataset into two subsets: a training+validation set with 13 cases and a test set with 4 cases. Importantly, the surgical cases included in the test set were completely unseen during the training process, ensuring an unbiased evaluation of the models' performance on new data. To further enhance the robustness of our models, we implemented an overlapping split strategy for the 13 cases in the training+validation set. 
Comparing the performance of ImageNet ViT, VideoMAE, and MVD on this challenging dataset, we aim to identify the most effective approach for SPR. The combination of pretrained models, fine-tuning, and a resiliently designed train-test split enables us to thoroughly evaluate the capabilities of these models and their potential for real-world application in surgical video analysis.

\section{Experiments}

\subsection{Dataset}

\begin{figure}[h]
    \centering
    \includegraphics[width=1\linewidth]{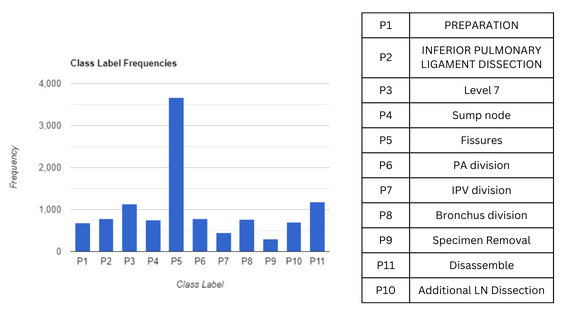}
    \caption{Frequency distribution of all the 11 phases listed in the table.}
    \label{fig:label-frequency}\vspace{-1em}
\end{figure}

The Surgical Phase dataset used in this work consists of 17 videos averaging 2.18 hours each and 11 classes of surgical phases, sourced from diverse patients at Massachusetts General Hospital(MGH). Fig.~\ref{fig:label-frequency} shows the frequency distribution of the class labels. To prepare the data for our experiments, we converted the video dataset into smaller clips of 10 seconds from a single surgical phase. We employ the sliding window method with a stride of 10 seconds to generate these clips, thereby guaranteeing that no data is repeated or overlapped. 

\vspace{-1em}

\subsection{Metrics}

To evaluate the performance of our models, we employed two widely used metrics: Top-1 Accuracy and Top-5 Accuracy. Top-1 Accuracy measures the percentage of video clips for which the model correctly predicts the surgical phase as its top prediction. On the other hand, Top-5 Accuracy considers a prediction as correct if the ground truth surgical phase is among the model's top 5 predictions. These metrics provide a comprehensive assessment of the models' ability to accurately classify surgical phases in video clips.

\subsection{Results}

\renewcommand{\arraystretch}{1.5}

\begin{table}[h]
    \centering
    \caption{Top-1 and Top-5 Accuracy results}
    \label{tab:results}
    \begin{tabular}{|c|c|c|}
        \hline
        \textbf{Model} & \textbf{Top-1 Accuracy} & \textbf{Top-5 Accuracy}\\
        \hline VideoMAE & 68.61 & 92.07\\
        \hline ImageNet ViT & 52.31 & 88.46\\
        \hline Ours(MVD) & \textbf{72.930} & \textbf{94.144}\\
        \hline
    \end{tabular}
\end{table}

Table \ref{tab:results} provides the Top-1 and Top-5 Accuracy for all the models that we experimented. MVD achieves a Top-1 Accuracy of 72.930\%, surpassing ImageNet ViT by a significant margin of 20.62 percentage points and VideoMAE by 4.32 percentage points. Similarly, MVD attains a Top-5 Accuracy of 94.144\%, demonstrating its superiority over ImageNet ViT and VideoMAE by 5.684 and 2.074 percentage points, respectively.

The superior performance of MVD can be attributed to its ability to effectively capture temporal dependencies and learn discriminative features from video data. By leveraging the power of masked video modeling and distillation techniques, MVD achieved highest Top-1 and Top-5 Accuracy which demonstrate its effectiveness in accurately identifying surgical phases.

\section{CONCLUSIONS AND FUTURE WORK}

In this work, we presented an analysis of Thoracic surgery video data to achieve Surgical Phase Recognition(SPR) using fine-tuned MVD model. MVD outperformed other models, making it a promising approach for SPR. In the near future we plan to generalize this model to other surgical procedures and use the predicted phases to summarize a surgery.

\addtolength{\textheight}{-12cm}   





\bibliographystyle{IEEEtran}
\bibliography{references}

\end{document}